\documentclass[conference]{IEEEtran}
\IEEEoverridecommandlockouts
\usepackage{amsmath,amssymb,amsfonts}
\usepackage{algorithmic}
\usepackage{graphicx}
\usepackage{textcomp}
\usepackage{xcolor}
\usepackage{float}
\usepackage{hyperref}
\hypersetup{
    colorlinks=true,
    linkcolor=blue,
    filecolor=magenta,      
    urlcolor=cyan,
}

\usepackage[round]{natbib}
\usepackage{hyperref}
\def\BibTeX{{\rm B\kern-.05em{\sc i\kern-.025em b}\kern-.08em
    T\kern-.1667em\lower.7ex\hbox{E}\kern-.125emX}}
\begin{document}

\title{Explaining model's visual explanations from its decision \\
}

\author{\IEEEauthorblockN{ Dipesh Tamboli}
\IEEEauthorblockA{
\textit{Indian Institute of Technology Bombay}, Mumbai, India \\
dipesh.tamboli@gmail.com}

}

\maketitle

\begin{abstract}
This document summarizes different visual explanations methods such as \{CAM, Grad-CAM, Localization using Multiple Instance Learning\} - \textit{Saliency based methods}, \{Saliency driven Class-Impressions, Muting pixels in input image\} - \textit{Adversarial methods} as well as \{Activation visualization, Convolution filter visualization\} - \textit{Feature-based methods}. We have also shown the results produced by different methods as well as a comparison between CAM, GradCAM and Guided Backpropagation.

\end{abstract}


\section{Introduction}
In recent years, Deep Neural Networks has shown impressive classification performance on a huge dataset such as Imagenet. Deep Learning regime empowers classifiers to extract distinct patterns of a given class from training data, which is the basis on which they generalize to unseen data. However, our understanding of how these models work or why they perform so well is not very clear. Before deploying these models on critical applications such as Medical Image Analysis, Road-Lane-Traffic light detection, etc., it is necessary to visualize the features considered to be important for making the decision.

There are several methods to generate visual explanations for a trained model. Saliency based methods first forward propagate the image and used the activation signal corresponding to the target class only for the backpropagation to the different layers of the network. These backpropagated gradient maps are combined differently to produce heatmaps to show the visualization. 

In the data-free saliency approach\citep{9190826}, noise is adversarially modified to maximize the confidence of the target class, which transforms it into the picture which model has in its memory. Muting the pixels in the input image and generating the heatmap from the confidence of the target class highlights the location where an important object is present in the image. 

Feature-based methods use weights of the model(specifically of convolution filter's) to conclude it. Zeiler and Fergus have shown that those filters resemble the Gabor filters. 

The organization of this document is briefly described here. In the following section, we described related literature in the field of Visualization of Deep Networks and showed their results. Section-\ref{sec:code} describes the project idea and explains the code. We conclude the paper with our analysis in Section-\ref{sec:results}. 


\section{Different Methods}
\subsection{Visualizing Live ConvNET Activations}
For visualizing what convolutional filters have learnt, \citep{yosinski2015understanding} has plotted the activation value of neurons directly in response to an image or video. As in convolutional network, filters are applied in a way that respects the underlying geometry of the input. But that's not the case for the fully connected network as the order of the units is irrelevant. Thus this method is useful only for visualizing the features of the convolutional network.

\begin{figure}[htbp]
\centerline{\includegraphics[width=8.0cm]{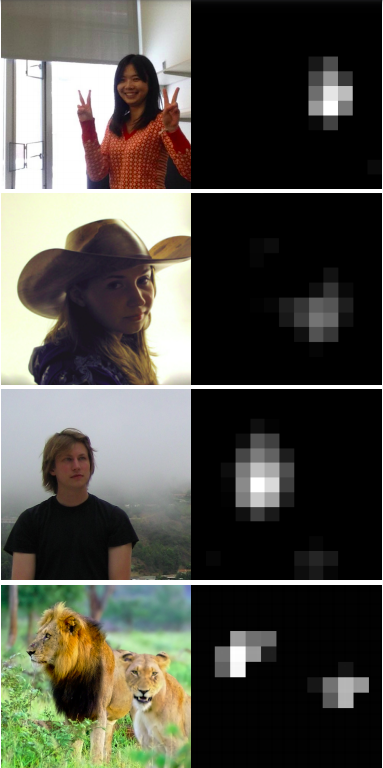}}
\caption{A view of the 13×13 activations of the 151st channel on the conv5 layer of a deep neural network trained on ImageNet, a dataset that does not contain a face class, but does contain many images with faces.}
\label{convnet_fig}
\end{figure}

In the Fig.\ref{convnet_fig}, the activation of a conv5 filter is shown which is trained on the Imagenet dataset. The point to be noted here is that the Imagenet dataset has no specific class for \textit{Face}, but still, the model can capture it as an important feature for the classification.

\subsection{Saliency-Driven Class Impressions}
In this input-agnostic data-free method \citep{9190826} has proposed a method of generating visualizations containing highly discriminative features learned by the network. They initially start with a noise image and iteratively update this using gradient ascent to maximize the logits of a given class. A set of transformations such as random rotation, scaling, RGB jittering and random cropping between iterations ensures that the generated images are robust to these transformations; a feature that natural images
typically possess. 

One of the key statistical properties that characterize a natural image is their spatial smoothness which was lacking in the above-generated images. Thus \textit{Total Variation Loss} is added to as a Natural Image Prior for making generated images more smooth.

\begin{figure}[htb]
\begin{minipage}[b]{0.48\linewidth}
  \centering
  \centerline{\includegraphics[width=4.0cm]{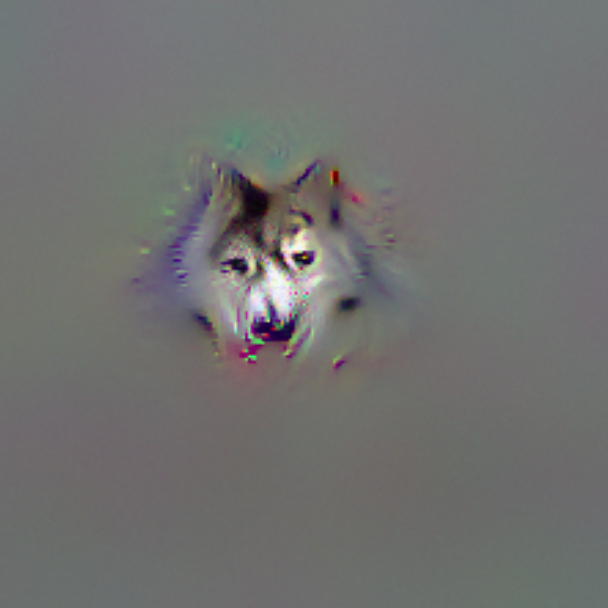}}
  \centerline{Fox}\medskip
\end{minipage}
\hfill
\begin{minipage}[b]{0.48\linewidth}
  \centering
  \centerline{\includegraphics[width=4.0cm]{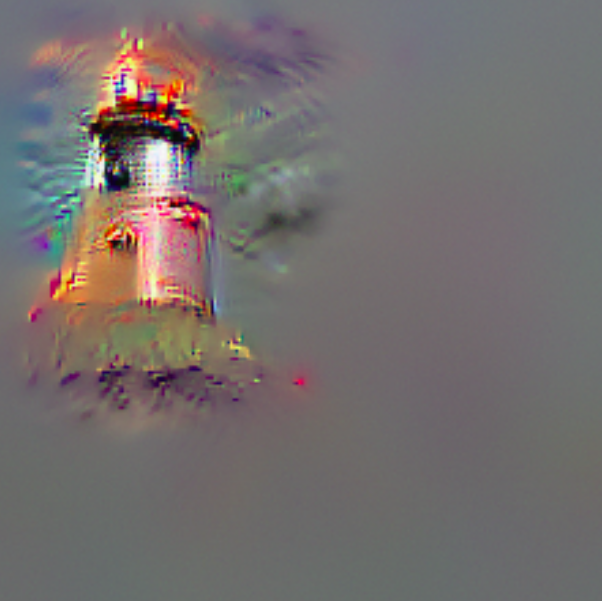}}
  \centerline{Watch-Tower}\medskip
\end{minipage}
\begin{minipage}[b]{0.48\linewidth}
  \centering
  \centerline{\includegraphics[width=4.0cm]{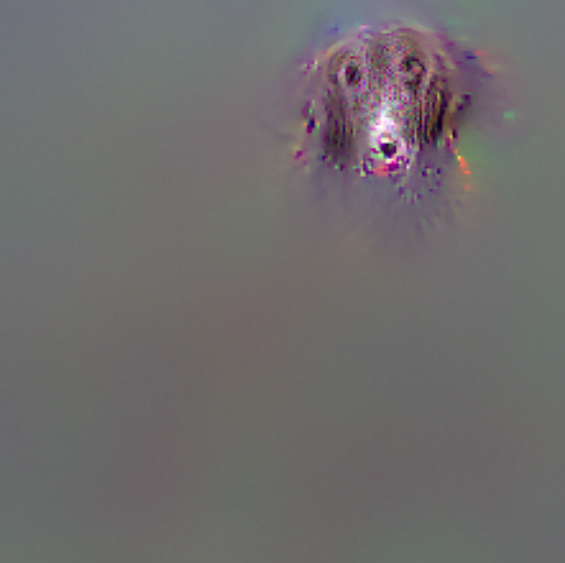}}
  \centerline{Dog}\medskip
\end{minipage}
\hfill
\begin{minipage}[b]{0.48\linewidth}
  \centering
  \centerline{\includegraphics[width=4.0cm]{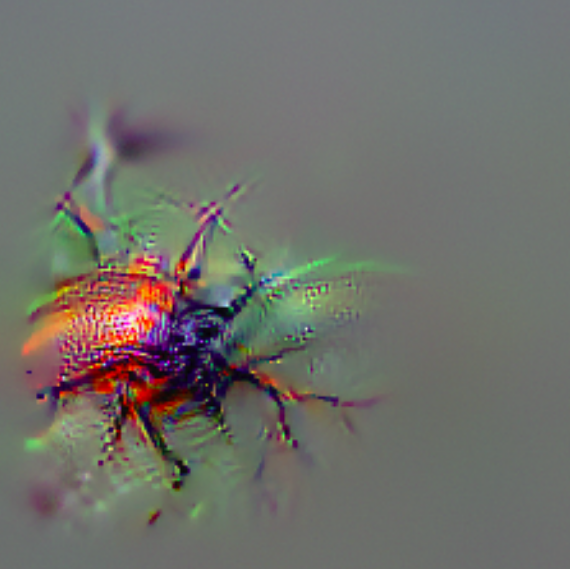}}
  \centerline{Spider}\medskip
\end{minipage}

\begin{minipage}[b]{0.48\linewidth}
  \centering
  \centerline{\includegraphics[width=4.0cm]{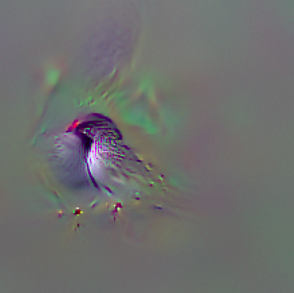}}
  \centerline{Bird}\medskip
\end{minipage}
\hfill
\begin{minipage}[b]{0.48\linewidth}
  \centering
  \centerline{\includegraphics[width=4.0cm]{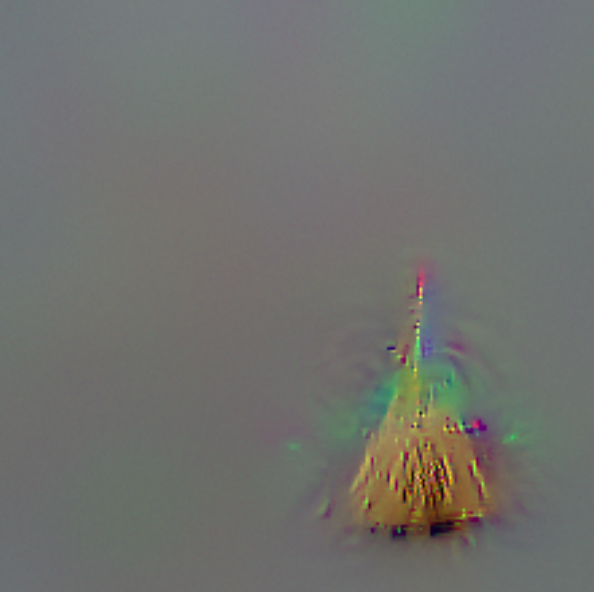}}
  \centerline{Broom}\medskip
\end{minipage}
\vspace{-0.5cm}
\caption{Saliency-Driven Class Impression from \citep{9190826}}

\label{CI_fig}
\end{figure}

Fig. \ref{CI_fig} generated Saliency-Driven Class Impressions represents what in-general model has learnt corresponding to a specific class. Surprisingly, the adversarially generated feature image also looks like an actual object and not just random noise.

\begin{figure*}[tb]
 \center
  \includegraphics[width=0.8\textwidth,height=8cm]{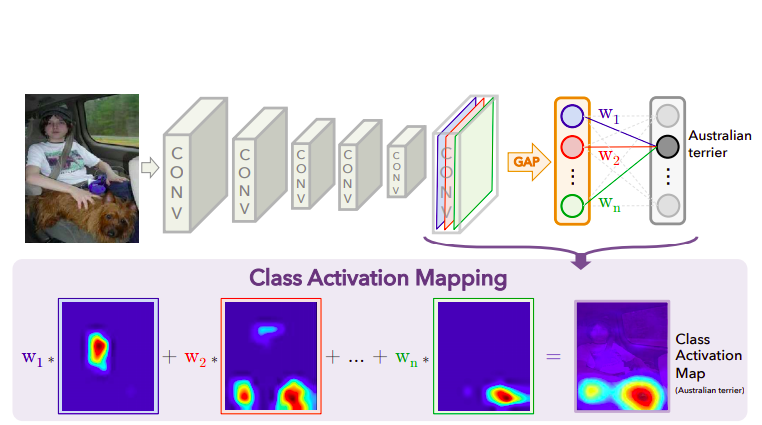}
  \caption{Working of CAM}
  \label{cam_arch}
\end{figure*}

\begin{figure*}[tb]
 \center
  \includegraphics[width=0.8\textwidth,height=8cm]{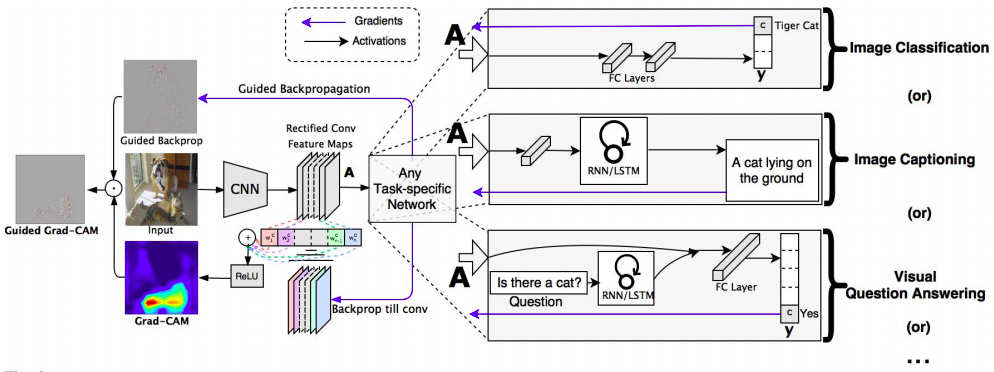}
  \caption{Working of Grad-CAM}
  \label{gradcam_arch}
\end{figure*}

\subsection{Localization using Multiple Instance Learning}
Gradient-based localization techniques do not produce good results on histopathology images as features are distributed over most of the part of the image. \citep{9019916} have shown that backpropagation methods use for generating some explanations are not useful in case of Histopathology images and thus propose a new method call attention-based multiple instance learning (A-MIL).

\begin{figure}[htbp]
\centerline{\includegraphics[width=8.0cm]{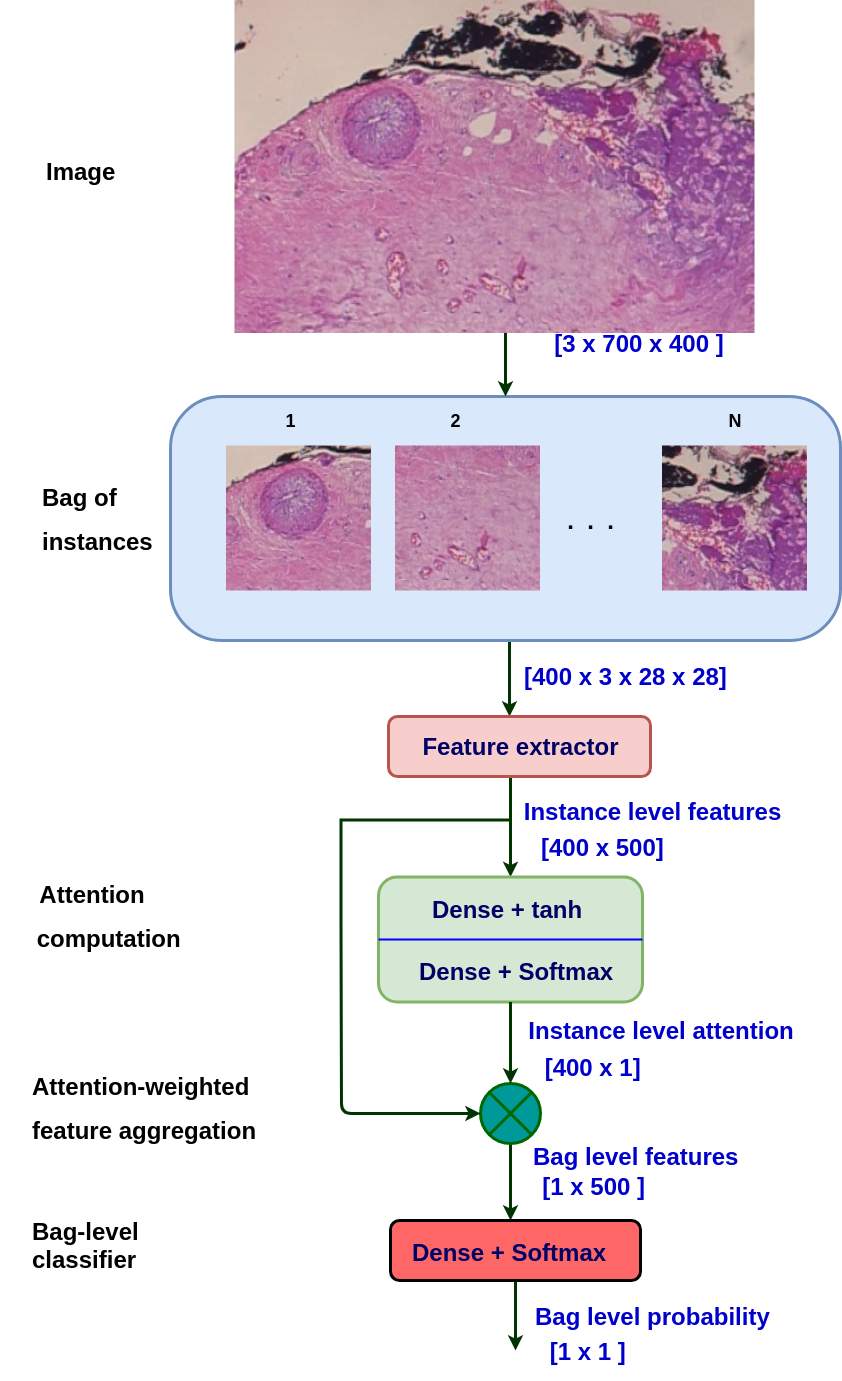}}
\caption{Architecture of the A-MIL method.}
\label{AMIL_arch}
\end{figure}

Fig.\ref{AMIL_arch}, an input image is shredded down to patches of the same size, which then passed to the classification network. This method provides the solution for a weakly supervised learning problem. Instance level pooling aggregates instance level features to obtain bag level features. This bag level features then passed through the network to get the confidence corresponding to each patch. This confidence is then used for highlighting the complete input image, thus brightening the patch which has high confidence and suppressing the portion which has low confidence.

\begin{figure}[htbp]
\centerline{\includegraphics[width=8.0cm]{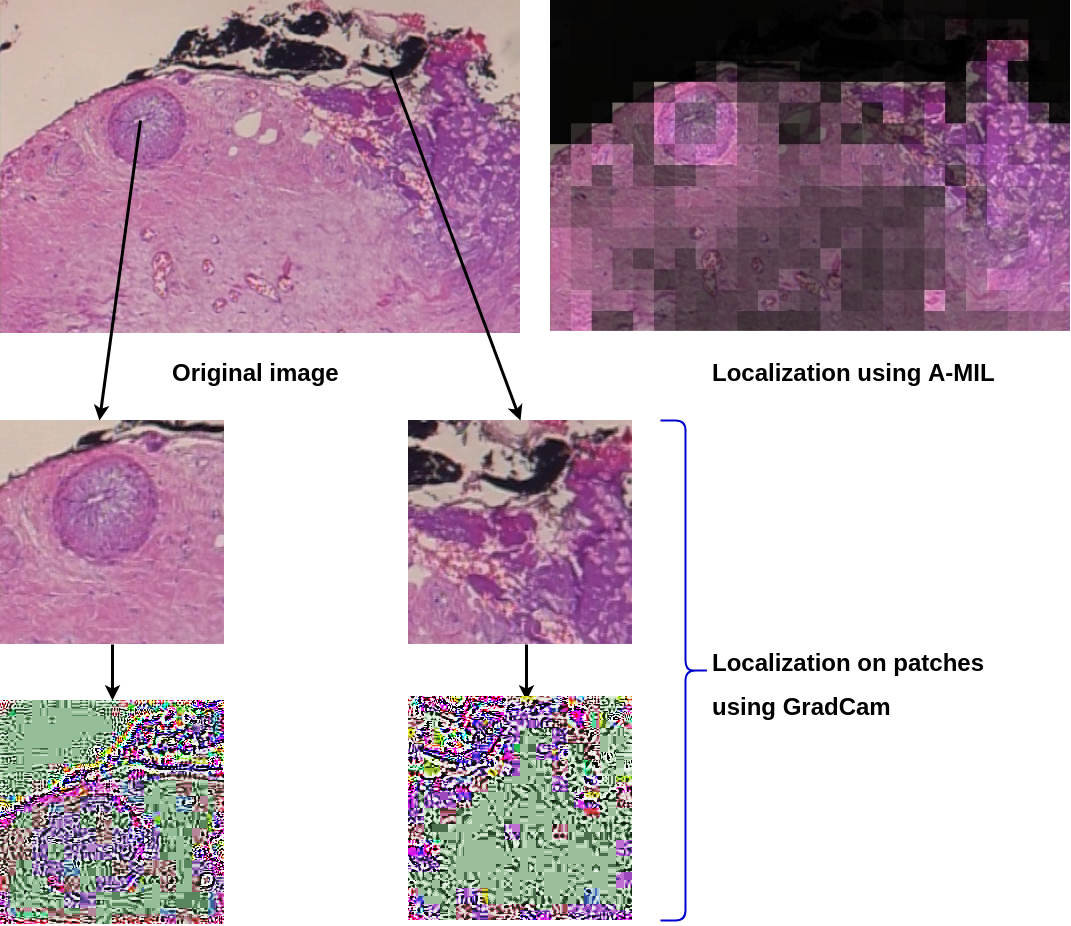}}
\caption{Visualization difference between explanations produced by GradCAM vs A-MIL}
\label{AMIL_result}
\end{figure}

Fig.\ref{AMIL_result} has shown the explanations provided by the popular GradCAM \citep{selvaraju2017grad} method and A-MIL. As claimed by the authors, GradCAM is highlighting in-general some random portion and not one which is important for the detection. However, the accuracy achieved by the GradCAM network was comparable with A-MIL.

\subsection{Gradient-weighted Class Activation Mapping (Grad-CAM)}
GradCAM \citep{selvaraju2017grad} uses the gradients of any target concept (say ‘dog’ in a classification network or a sequence of words in the captioning network) flowing into the final convolutional layer to produce a coarse localization map highlighting the important regions in the image for predicting the concept. 

As convolutional layers naturally retain the spatial information which gets lost in fully connected layers, we expect the last convolutional layers(also called as \textit{Rectified Conv Feature Maps}) to have the best compromise between high-level semantics and detailed spatial information.

\subsubsection*{How Grad-CAM is different from CAM}
Class Activation Mappings(CAM, \cite{Zhou_2016_CVPR}) Produces a localization map for an image classification CNN with a specific kind of architecture where global average pooled convolutional feature maps are fed directly into softmax. These feature maps are then spatially pooled using Global Average Pooling (GAP) and linearly transformed to produce a score for each class.

Here the limitation for the CAM comes from the fact that CAM is applicable only on the architectures where final fully connected layers are not there. 
Fig.\ref{cam_arch} shows the procedure to get the final weight vector.

On the contrary, Grad-CAM(Fig. \ref{gradcam_arch}) works with all type of architecture, even where fully connected layers are used. 

Given an image and a class of interest (e.g., ‘tiger cat’ or any other type of differentiable output) as input, GradCAM forward propagate the image through the CNN part of the model and then through task-specific computations to obtain a raw score for the category. The gradients are set to zero for all classes except the desired class (tiger cat), which is set to 1. This signal is then backpropagated to the rectified convolutional feature maps of interest, which is combined to compute the coarse Grad-CAM localization (blue heatmap) which represents where the model has to look to make the particular decision. Finally, it multiplies the heatmap pointwise with guided backpropagation to get Guided Grad-CAM visualizations which are both high-resolution and concept-specific. 

\section{Implementation and code}
\label{sec:code}
\subsection{Project}
\begin{itemize}
    \item Doing a thorough literature review of the existing visualization methods used for locating the important portion in the image responsible for the model's decision
    \item Implementing CAM and Grad-CAM along with the Guided-backprop for a \textbf{multi-label} classification setup.
    \item Preparation of a Google-Colab interactive notebook \citep{Dipeshtamboli} where you can directly upload an image and get corresponding GradCAM visualization along with the top 5 class predictions. This ready-to-use implementation is ready for the model trained on COCO dataset as well as Imagenet dataset.
\end{itemize}

\subsection{Code: How to use it}
This \href{https://colab.research.google.com/github/Dipeshtamboli/Interactive-GradCAM/blob/master/Interactive_GradCAM_ImageNet.ipynb#scrollTo=1eWQO989JiaR}{Google-Colab notebook} is self-sufficient to run the GradCAM inference on any image corresponding to the network either trained on Imagenet dataset or \href{https://colab.research.google.com/drive/1sD2sWq7rNz3CIJvcPsX-G17MV9EqV2g5}{COCO} dataset. 

GradCAM requires a network and corresponding trained weights to get an inference. As I have implemented GradCAM for multi-label classification, I needed to change the architecture, and thus we cannot use PyTorch's pre-trained models. Thus I am hosting the weights corresponding to the modified network from my GitHub repository\citep{Dipeshtamboli} and downloading it to the Colab notebook while running. Also, some samples images are present in the repository which is by-default get downloaded in the Colab environment for the testing purpose.

\section{Results, Conclusions and Error Analysis}
\label{sec:results}

\begin{figure*}[tb]
 \center
  \includegraphics[width=\textwidth]{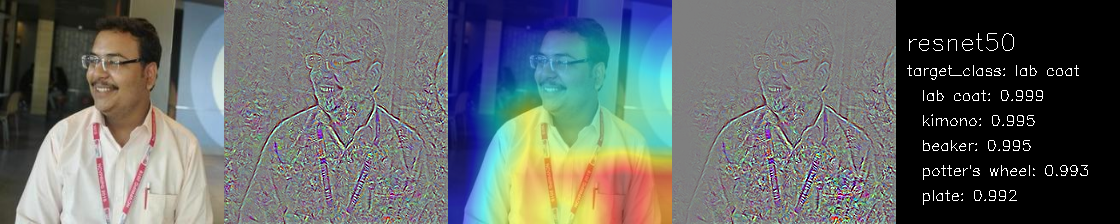}
  \caption{Test image of myself, detected as \textbf{Lab coat} class}
  \label{dipesh_labcoat}
\end{figure*}    
\begin{figure*}[tb]
 \center
  \includegraphics[width=\textwidth]{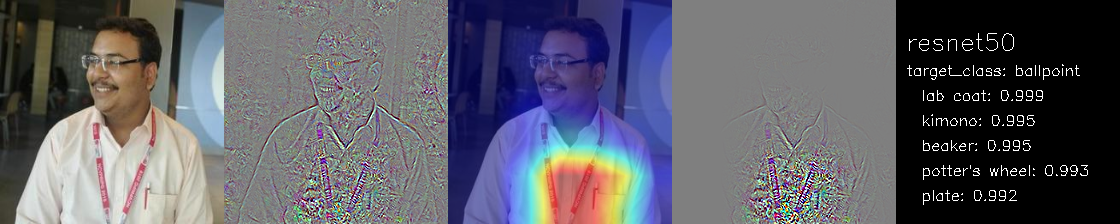}
  \caption{Target class: \textbf{Ball point pen}, can see the focus shifted to the pocket from \ref{dipesh_labcoat}}
  \label{dipesh_ballpen}
\end{figure*}    

\begin{figure*}[tb]
 \center
  \includegraphics[width=\textwidth]{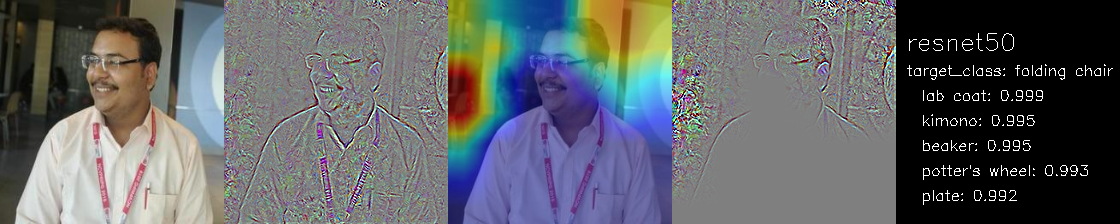}
  \caption{Target class: \textbf{Folding Chair}, the focus has shifted to the tiny chair in the background}
  \label{dipesh_chair}
\end{figure*}    

For the Fig. \ref{dipesh_labcoat}, we check the activation for the predicted class of the network, which turns out to be a lab coat. Also, the region specified by the Grad-CAM was consistent with its prediction.

After that, in Fig. \ref{dipesh_ballpen}, we tested the Grad-CAM activation corresponding to a target class: \textbf{Ball Pen}, and we found out that the highlighted region shifts towards the pocket of the shirt where the actual Ballpen is present. After in fig. \ref{dipesh_chair}, when we specify our target class to be a \textbf{Folding chair}, Grad-CAM highlighted the chair present in the background of the input image.

\begin{figure*}[tb]
 \center
  \includegraphics[width=\textwidth]{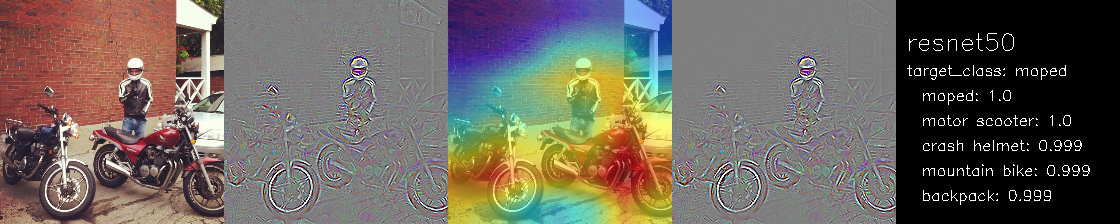}
  \caption{Target class: \textbf{Moped}, highlighting the 2 mopeds along with the person}
  \label{moped_rider}
\end{figure*}

\begin{figure*}[tb]
 \center
  \includegraphics[width=\textwidth]{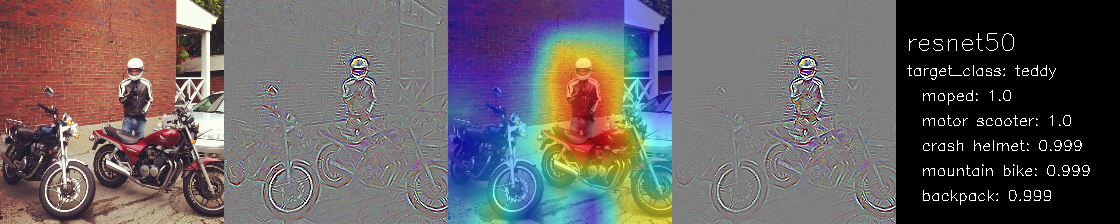}
  \caption{Target class: \textbf{Teddy}, though the person is not a teddy but closest to one, GradCAM is highlighting the person as teddy}
  \label{teddy_rider}
\end{figure*}

\begin{figure*}[tb]
 \center
  \includegraphics[width=\textwidth]{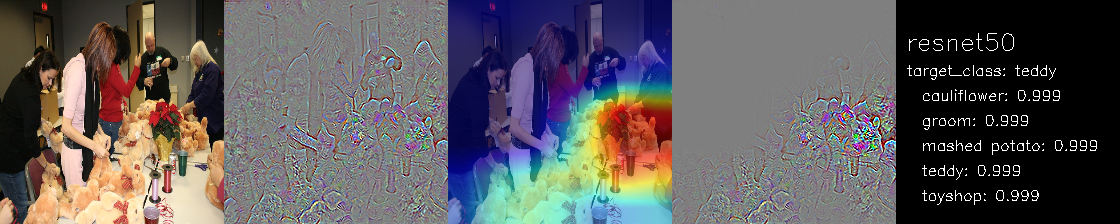}
  \caption{Target class: \textbf{Teddy}, here, GradCAM can distinguish between the person and teddy class}
  \label{party_teddy}
\end{figure*}

Fig. \ref{moped_rider}, when the target class is \textbf{moped}, GradCAM is highlighting moped along with the person present in the frame. When we change our target class to \textbf{teddy} in fig. \ref{moped_rider}, although teddy is not present in the image but person is the closest object to it, GradCAM is highlighting the person in the frame. But when we input an image where teddy and person, both classes are present(fig. \ref{party_teddy}), GradCAM highlights only teddies present on the table and not the people around it.

\newpage
\textbf{Some critical observations:}
\begin{itemize}
    \item Although GradCAM can detect the objects(pen, chair) properly, but still, the probability for those classes are not in the top5.
    \item As this is a multi-label classification setup, the sum of probabilities of all the classes does not sum up to the one, and thus multiple importance regions are visible.
    \item Fig. \ref{teddy_rider}, GradCAM is highlighting the person for the class teddy as it is the closest to the teddy class. Still, in fig. \ref{party_teddy}, both the teddy and person objects are present, GradCAM is properly locating only teddies present on the table and not the people standing around it. This ensures that although the model thinks a person to be the nearest class to the teddy, it is also able to discriminate between both.
\end{itemize}

\vspace{12pt}

\bibliography{citations}

\end{document}